%% file: main.tex
\pdfoutput=1 
\documentclass[11pt,a4paper]{article}
\usepackage[hyperref]{acl2020}
\usepackage{times}
\usepackage{latexsym}

\usepackage{booktabs}
\usepackage{microtype}
\usepackage{amsmath}
\usepackage{amssymb}
\usepackage{multirow}

\aclfinalcopy 

\title{
Semi-Autoregressive Training Improves Mask-Predict Decoding
}

\author{
Marjan Ghazvininejad \\
\\
\And 
Omer Levy \\
\\
Facebook AI Research 
\And
Luke Zettlemoyer \\
\\
}

\date{}

\begin{document}
\maketitle

\input{00-abstract.tex}
\input{01-intro.tex}
\input{02-background.tex}
\input{03-method.tex}

\input{04-experiments.tex}

\input{05-related.tex}
\input{06-conclusion.tex}

\bibliographystyle{acl_natbib}
\bibliography{references}

\end{document}

%% file: 00-abstract.tex
\begin{abstract}
The recently proposed mask-predict decoding algorithm has narrowed the performance gap between semi-autoregressive machine translation models and the traditional left-to-right approach.
We introduce a new training method for conditional masked language models, SMART, which mimics the semi-autoregressive behavior of mask-predict, producing training examples that contain model predictions as part of their inputs. Models trained with SMART produce higher-quality translations when using mask-predict decoding, effectively closing the remaining performance gap with fully autoregressive models.
\end{abstract}

%% file: 01-intro.tex
\section{Introduction}

While mainstream approaches to machine translation sequentially generate a translation token by token, recent advances in non-autoregressive \cite{gu2017,libovick2018,sun2019} and semi-autoregressive decoding \cite{lee2018,stern2019,gu2019} have produced increasingly viable alternatives, which can decode substantially faster, with some cost to performance.
One such approach, mask-predict \cite{ghazvininejad2019}, 
repeatedly predicts the entire target sequence in parallel,
conditioned on the most confident word predictions from the previous iteration.
The underlying model, a conditional masked language model, is trained by masking part of the (gold) target sequence and predicting the missing tokens.
During training, all observed (unmasked) tokens come from the ground truth data. However, at inference time, the observed tokens are high-confidence model predictions, creating a discrepancy that can hurt performance in practice.

To remedy this problem, we introduce SMART (\textbf{S}e\textbf{m}i-\textbf{A}uto\textbf{r}egressive \textbf{T}raining), a new training process for conditional masked language models that better matches the semi-autoregressive nature of the mask-predict decoding algorithm.
We first create training examples by starting with the gold target sequence and masking a subset of its tokens, just like the original training process. We then use the current model to predict the sequence from the partially-observed input, and  mask a different subset of tokens to create the training example's input.
The model is then trained to predict the gold target sequence based on this partially-observed prediction-based input, as well as the source sequence (see Figure~\ref{fig:example}), allowing it to better correct mistakes made during the early iterations of the mask-predict decoding loop.

SMART improves the performance of mask-predict decoding by 0.5 to 1.0 BLEU, effectively closing the gap with fully autoregressive models.
For example, in the WMT'14 EN-DE benchmark, we arrive at a BLEU score of 27.65, just under the  27.75 achieved by a strong autoregressive baseline.
This result implies that the of mask-predict decoding is not only a fast alternative to autoregressive beam search, but also an accurate one.

%% file: 02-background.tex
\begin{figure*}[th]
    \centering
    \small
    \begin{tabular}{@{}lccccccccccc@{}}
        \multicolumn{12}{c}{\textbf{Example Generation Steps}} \\
        \toprule
        $Y^{\text{gold}}$ & The & hotel & is & an & ideal & choice & for  & business & and & leisure & trips \\
        $Y^{\text{gold}}_{\text{obs}}$  & --- & --- & is & --- & --- & --- & for & --- & --- & leisure & --- \\  
        $Y^{\text{pred}}$ & The & hotel & is & an & choice & choice & for & business & and & leisure & travellers \\
        $Y^{\text{pred}}_{\text{obs}}$ &  The & --- & is & an & choice & choice & for & --- & --- & --- & travellers \\
        \bottomrule
        \\
        \multicolumn{12}{c}{\textbf{Final Training Example}} \\
        \toprule
        $X$ &  Das & Hotel & ist & eine & ideale & Wahl & für & Geschäfts- & und & Urlaubs & @@reisen \\
        $Y^{\text{pred}}_{\text{obs}}$ &  The & --- & is & an & choice & choice & for & --- & --- & --- & travellers \\
        \midrule
        $Y^{\text{gold}}$ &  The & hotel & is & an & ideal & choice & for  & business & and & leisure & trips \\
        \bottomrule
    \end{tabular}
    \caption{An illustration of how SMART generates new training examples. We start with the gold target sequence $Y^{\text{gold}}$ and randomly mask some of its tokens, use the partially-observed gold sequence $Y^{\text{gold}}_{\text{obs}}$ to predict the entire translation $Y^{\text{pred}}$, and then mask a random subset of tokens again. The resulting sequence is used as the model's input during optimization, alongside the source $X$, when training to predict the original gold sequence $Y^{\text{gold}}$.}
    \label{fig:example}
\end{figure*}

\section{Background: Mask-Predict\footnote{For further detail, see \cite{ghazvininejad2019}.}}
\label{sec:background}

\paragraph{Conditional Masked Language Models}
A conditional masked language model (CMLM) takes a source sequence $X$ and a partially-observed target sequence $Y_{\text{obs}}$ as input. 
It predicts the probabilities of the masked (unobserved) target sequence tokens $Y_{\text{mask}}$, assuming conditional independence between them (given the inputs). 

Since each target token $y \in Y$ is either observed or masked,
the predictions are effectively conditioned on the target sequence length $N$ as well, which must be predicted separately by the model.

\paragraph{Mask-Predict Decoding}
Mask-predict generates the entire target sequence in a preset number of decoding iterations $T$.
Given the predicted target sequence length $N$, decoding starts with a fully-masked target sequence.\footnote{In practice, the algorithm uses multiple length candidates, decodes each in parallel, and selects the best (highest-probability) result. Considering multiple length candidates is somewhat analogous to beam search in autoregressive decoding.}
The model then predicts the entire sequence in parallel, setting each token $y_i$ with its most probable assignment $w$ ($\arg\max_w P(y_i = w)$).

For each iteration $2 \leq t \leq T$, the algorithm performs a \emph{mask} step, in which the tokens with the lowest probabilities are replaced with a special mask token.\footnote{The number of masked tokens gradually shrinks with $t$.} 
This is followed by a \emph{predict} step, where the model predicts the masked tokens while conditioning on the observed high-confidence predictions from the previous iterations.

\paragraph{Non-Autoregressive Training}
The original training process for CMLMs takes the gold target sequence and masks out $k$ random tokens, where $k \sim \text{Uniform} (1, N)$.
The model then predicts only the masked tokens while conditioning on the observed target tokens, which are always correct.
Training optimizes the cross-entropy between the predictions and the correct values of the masked tokens.
We call this process NART (\textbf{N}on-\textbf{A}uto\textbf{r}egressive \textbf{T}raining) because it only uses gold data as its inputs, and does not condition on model predictions.

%% file: 03-method.tex
\section{Semi-Autoregressive Training}
\label{sec:method}

The non-autoregressive training process of \citet{ghazvininejad2019}, NART, creates training examples where all the observed tokens are correct -- the right word type in the right position.
This assumption does not hold for mask-predict decoding, since the observed tokens (high-confidence predictions from previous iterations) are \emph{not} always correct.
We introduce an improved training process for CMLMs that better reflects the semi-autoregressive nature of mask-predict decoding by creating training examples from predicted target sequences, not gold ones. We name this training procedure SMART (\textbf{S}e\textbf{m}i-\textbf{A}uto\textbf{r}egressive \textbf{T}raining).

Like NART, we start with a gold target sequence $Y^{\text{gold}}$ and randomly mask $k_{\text{gold}}$ tokens, where $k_{\text{gold}}$ is sampled uniformly from 1 to $N$ (the target's length).
The CMLM then predicts the entire sequence, including the observed tokens, creating a new sequence $Y^{\text{pred}}$ from the most probable assignments.
We repeat the masking process, but with different random values, to construct the final training example; i.e. we sample $k_{\text{pred}} \sim \text{Uniform} (1, N)$ and randomly mask $k_{\text{pred}}$ tokens from $Y^{\text{pred}}$ to create the partially-observed target sequence $Y^{\text{pred}}_{\text{obs}}$. 
Figure~\ref{fig:example} illustrates this process.\footnote{We perform a double forward pass only when creating training examples. During inference, each mask-predict iteration includes only a single forward pass in the \emph{predict} step.}

The observed portion of $Y$ may contain incorrect observations because it is based on predictions ($Y_{\text{pred}}$).
Therefore, we optimize the cross entropy for predicting all tokens, not only the masked ones. 
This change allows models trained with SMART to fix incorrect observations during prediction, and can be integrated into the mask-predict algorithm by modifying the \emph{predict} step:
instead of predicting just the masked tokens, predict \emph{every} target token, and update those tokens whose predictions differ from the input.

%% file: 04-experiments.tex
\begin{table*}[t!]
\centering
\small
\begin{tabular}{lrcccc}
\toprule
\multirow{2}{*}{\textbf{Training Mode}} & \textbf{Decoding} & \multicolumn{2}{c}{\textbf{WMT'14}} & \multicolumn{2}{c}{\textbf{WMT'17}} \\
& \textbf{Iterations} & \textbf{EN-DE} & \textbf{DE-EN} & \textbf{EN-ZH} & \textbf{ZH-EN} \\
\midrule
\textit{NART}
& 1~~~~~~ & 18.05 & 21.83 & \textbf{24.23} & \textbf{13.64} \\
\textit{SMART}
& 1~~~~~~ & \textbf{18.58} & \textbf{23.77} & 24.15 & 13.51 \\
\midrule
\textit{NART}
& 4~~~~~~ & 25.94 & 29.90 &  32.63 & 21.90  \\
\textit{SMART}
& 4~~~~~~ & \textbf{27.03} & \textbf{30.87} &  \textbf{33.37} & \textbf{22.61}\\
\midrule
\textit{NART}
& 10~~~~~~ & 27.03 & 30.53 & 33.19 &  23.21 \\
\textit{SMART}
& 10~~~~~~ & \textbf{27.65} & \textbf{31.27} & \textbf{34.06} &  \textbf{23.78}\\
\bottomrule
\end{tabular}
\caption{The performance (test set BLEU) of semi-autoregressive training (SMART), compared to the original non-autoregressive training for CMLMs (NART). All models are decoded with mask-predict.}
\label{tab:results_maskpredict}
\end{table*}

\begin{table*}[t!]
\centering
\small
\begin{tabular}{lrcccc}
\toprule
\multirow{2}{*}{\textbf{Model}} & \textbf{Decoding} & \multicolumn{2}{c}{\textbf{WMT'14}} & \multicolumn{2}{c}{\textbf{WMT'17}} \\
&  \textbf{Iterations}  & \textbf{EN-DE} & \textbf{DE-EN} & \textbf{EN-ZH} & \textbf{ZH-EN} \\
\midrule
\textit{Autoregressive Transformer with Beam Search} 
& $N$~~~~~~ & 27.61 & 31.38 & 34.31 &  23.65 \\
\textit{~~~~+ Knowledge Distillation} 
& $N$~~~~~~ & \textbf{27.75} & 31.30 & \textbf{34.38} &  23.91 \\
\midrule 
\multirow{2}{*}{\textit{SMART CMLM with Mask-Predict}}
& 10~~~~~~ & 27.65 & 31.27 & 34.06 &  23.78\\
& $N$~~~~~~ & 27.64 & \textbf{31.44} & 34.10 & \textbf{24.12} \\
\bottomrule
\end{tabular}
\caption{The performance (test set BLEU) of semi-autoregressive training (SMART), compared to the standard (sequential) transformer. Length beam, beam size and length penalty is tuned for each model on validation set. }
\label{tab:results}
\end{table*}

\section{Experiments}

We demonstrate, over 4 benchmarks, that replacing the original CMLM training process with SMART produces higher quality translations when decoding with mask-predict.
Moreover, we show that our new approach closes the performance gap between semi-autoregressive and fully autoregressive machine translation.
Finally, we conduct an ablation study and analyze how SMART balances between easy and hard training examples.

\subsection{Setup}

We evaluate on two machine translation datasets, in both directions (four benchmarks overall): WMT'14 English-German (4.5M sentence pairs), and WMT'17 English-Chinese (20M sentence pairs).
The datasets are tokenized into subword units using BPE \cite{sennrich2016}.
We use the same preprocessed data as \citet{vaswani2017} and \citet{wu2019pay} for WMT'14 EN-DE and WMT'17 EN-ZH respectively.
We evaluate performance with BLEU \cite{papineni2002} for all language pairs, except from English to Chinese, where we use SacreBLEU \cite{post2018}.\footnote{SacreBLEU hash: BLEU+case.mixed+lang.en-zh +numrefs.1+smooth.exp+test.wmt17+tok.zh+version.1.3.7}

We implement our experiments based on the code of mask-predict \cite{ghazvininejad2019}, which uses the standard model and optimization hyperparameters for transformers in the base configuration \cite{vaswani2017}: 512 model dimensions, 2048 hidden dimensions, model averaging, etc.
We also follow the standard practice of knowledge distillation  \cite{gu2017,ghazvininejad2019,zhou2019understanding} in the non-autoregressive machine translation literature, and train both our model and the baselines on translations produced by a large autoregressive transformer model.
For autoregressive decoding, we tune the beam size ($b \in \{ 1 \ldots 7 \}$) and length penalty on the development set, and similarly tune the number of length candidates ($\ell \in \{ 1 \ldots 7 \}$) for mask-predict decoding.

\subsection{Results}

We first compare SMART to the original CMLM training process (NART). Table~\ref{tab:results_maskpredict} shows that SMART typically produces better models, with an average gain of 0.71 BLEU.
Even with a single decoding iteration (the purely non-autoregressive scenario), SMART produces better models in WMT'14 and falls short of the baseline by a slim margin in WMT'17 (0.08 and 0.13 BLEU).\footnote{We show the NART numbers reported by \citet{ghazvininejad2019}, where $\ell = 5$ length candidates were used. For fair comparison, we decoded the NART models while tuning the number of length candidates $\ell$ on the development set, but observed only minor deviations from the $\ell = 5$ setting.}


We also compare between SMART-trained CMLMs with mask-predict decoding and autoregressive transformers with beam search.
Table~\ref{tab:results} shows that a constant number of decoding steps (10) brings our semi-autoregressive approach very close to the autoregressive baseline. 
With the exception of English to Chinese, the performance differences are within the typical random seed variance. 
Increasing the number of mask-predict iterations to $N$ yields even more balanced results; in two of the four benchmarks, the small performance margins are actually in favor of our semi-autoregressive approach.


\subsection{Ablation Study}

We consider several variations of our proposed method to quantify the effect of each component. To prevent overfitting, we evaluate on the development set using $\ell = 5$ length candidates.

\paragraph{Repredicting All Tokens}
Besides SMART, we also augment the mask-predict algorithm to predict all tokens -- not only the masked ones -- during the \emph{predict} step (Section~\ref{sec:method}).
Table~\ref{tab:results_all_tokens} compares this new version of mask-predict to the original.
We find that predicting all tokens increases performance by 0.40 BLEU on average when using 4 decoding iterations.
With 10 decoding iterations, the gains shrink to around 0.08, but are still consistently positive.

\begin{table}[t!]
\centering
\small
\begin{tabular}{lrll}
\toprule
\multirow{2}{*}{\textbf{Predicted Tokens}} & \textbf{Decoding} & \multicolumn{2}{c}{\textbf{WMT'14}} \\
& \textbf{Iterations} & \textbf{EN-DE} & \textbf{DE-EN}\\
\midrule
\textit{Masked Tokens}
& 4~~~~~~ & 25.18 & 29.61 \\
\textit{All Tokens}
& 4~~~~~~ & \textbf{25.61}&  \textbf{29.98} \\
\midrule
\textit{Masked Tokens}
& 10~~~~~~ & 26.06 &  30.29 \\
\textit{All Tokens}
& 10~~~~~~ & \textbf{26.14}& \textbf{30.37} \\
\bottomrule
\end{tabular}
\caption{The performance (development set BLEU) of SMART-trained models with two flavors of mask-predict: predicting only masked tokens (original version), and predicting all tokens at each iteration.}
\label{tab:results_all_tokens}
\end{table}

\begin{table}[t!]
\centering
\small
\begin{tabular}{cccc}
\toprule
\multirow{2}{*}{\textbf{Forward Passes}} & \multicolumn{3}{c}{\textbf{Decoding Iterations}}   \\
& \textbf{1} &  \textbf{4} &  \textbf{10}  \\
\midrule
2 & \textbf{24.24} & \textbf{29.98} & \textbf{30.37}\\
3 & 23.74 & 29.89 & 30.28\\
4 & 23.81 & 39.66 &30.10\\
\bottomrule
\end{tabular}
\caption{Increasing the number of forward passes used to produce each training example in SMART can negatively effect the resulting model. Performance measured on WMT'14 DE-EN (development set BLEU).}
\label{tab:training_pass}
\end{table}

\begin{table}[t!]
\centering
\small
\begin{tabular}{cccc}
\toprule
\multirow{2}{*}{\textbf{Cross-Entropy Loss}} & \multicolumn{3}{c}{\textbf{Decoding Iterations}}   \\
 &\textbf{1} &  \textbf{4} &  \textbf{10}  \\
\midrule
\textit{1st Pass + 2nd Pass} & 23.89 & 29.78  & 30.05 \\
\textit{Only 2nd Pass} & \textbf{24.24} & \textbf{29.98} & \textbf{30.37}\\
\bottomrule
\end{tabular}
\caption{Using gradients from the first forward pass in SMART can negatively effect the resulting model. Performance measured on WMT'14 DE-EN (development set BLEU).}
\label{tab:all_losses}
\end{table} 

\begin{table}[t!]
\centering
\small
\begin{tabular}{cccc}
\toprule
\textbf{Gold Mask Ratio} & \multicolumn{3}{c}{\textbf{Decoding Iterations}}   \\
\textbf{($k_\text{gold} / N$)} & \textbf{1} & \textbf{4} & \textbf{10} \\
\midrule
~~~~0\% & 23.05 & 29.30 & 29.69\\
~~25\%& 23.19 & 29.41 & 29.84 \\
~~50\% & 23.04 & \textbf{29.99} & 30.15\\
~~75\% & \textbf{23.40} & 29.87 & \textbf{30.36}\\
100\% &16.78 & 18.44 & 18.62\\
\midrule
\textit{Uniform} & 24.24 & 29.98 & 30.37 \\
\bottomrule
\end{tabular}
\caption{The effect of the gold masking ratio (as a proxy of training example difficulty) on performance, measured on WMT'14 DE-EN (development set BLEU).}
\label{tab:mask_ratio}
\end{table}

\paragraph{Multi-Iteration SMART}
\citet{lee2018} also proposed a semi-autoregressive training regime, in which the training process imitated the iterative refinement decoding algorithm. They use four decoding iterations during training, while accumulating the gradients from every model invocation. We try to apply the same ideas to SMART, but find that they do not improve our method.

We first consider creating our training examples by performing multiple mask-predict iterations during training, instead of just two.
Table~\ref{tab:training_pass} shows that training on examples created by three or four forward passes of the model yields slightly (but consistently) worse results.

We also experiment with applying the cross-entropy loss after each forward pass (instead of just the last one).
Table~\ref{tab:all_losses} reveals that using these gradients produces slightly weaker models, suggesting that using only the examples produced by the latter forward pass provides the model with a better training signal.

\paragraph{Difficulty Analysis}

SMART produces training examples from model predictions conditioned on partially-observed gold data ($Y^\text{gold}_\text{obs}$). Intuitively, the amount of masked gold data will affect the difficulty of said example. 
When 0\% of the gold tokens are masked, the model will likely just copy its input ($Y^\text{pred} = Y^\text{gold}_\text{obs} = Y^\text{gold}$), and produce easier training examples, effectively reducing SMART to NART.
When 100\% of the gold tokens are masked, the training example will be entirely prediction-based, posing a significantly harder challenge for the model.

To explore the effect of training example difficulty on performance, we replace the uniformly distributed number of masks $k_\text{gold}$ with different fixed ratios.
Table~\ref{tab:mask_ratio} shows training with harder examples (50\% to 75\% gold mask ratio) improves performance, but that training with inputs that are not based on ``a grain of truth'' (100\% gold mask ratio) is not conducive to a successful learning process.
By sampling $k_\text{gold}$ from a uniform distribution, SMART provides training examples from a broad spectrum of difficulties.

%% file: 05-related.tex
\section{Related Work}

SMART was inspired by the iterative refinement model of \citet{lee2018}, who also used a semi-autoregressive training method. While Lee et al. seed their model inputs with artificial noise during training, the only source of noise in SMART is the model predictions.

Other semi-autoregressive models have also been able to close the performance gap with beam search decoded autoregressive models.
\citet{shu2019latent} demonstrate how a latent-variable approach can outperform the autoregressive baseline on Japanese to English translation, but still observe a significant performance gap on WMT'14 EN-DE.
Others have introduced insertion operators \cite{stern2019};
for example, the Levenshtein transformer \cite{gu2019levenshtein} allows for both insertions and deletions, achieving equal-quality translations with a smaller number of decoding iterations.
SMART achieves a similar result with a simple approach that requires neither latent variables nor insertions.

%% file: 06-conclusion.tex
\section{Conclusion}

We introduced SMART (\textbf{S}e\textbf{m}i-\textbf{A}uto\textbf{r}egressive \textbf{T}raining), a new training process for conditional masked language models that better matches the semi-autoregressive nature of the mask-predict decoding algorithm. SMART training produces models that are competitive with mainstream autoregressive models in terms of performance, while retaining the benefits of fast parallel decoding.